\documentclass{article} 
\usepackage{iclr2025_conference,times}

\usepackage[graphicx]{realboxes}
\usepackage{float}
\usepackage{amsmath}

\usepackage{multirow}
\usepackage{pifont}
\usepackage{tabularx}

\iclrfinaltrue

\usepackage{algorithm}
\usepackage{algpseudocode}

\usepackage{booktabs}       

\usepackage{amsmath,amsfonts,bm}

\def\eqref#1{equation~\ref{#1}}

\def\1{\bm{1}}

\DeclareMathAlphabet{\mathsfit}{\encodingdefault}{\sfdefault}{m}{sl}
\SetMathAlphabet{\mathsfit}{bold}{\encodingdefault}{\sfdefault}{bx}{n}

\def\wrt{\mbox{w.r.t.}}

\usepackage{hyperref}
\usepackage{url}

\newcommand{\shulian}[1]{\textcolor{black}{#1}}

\newcommand{\yong}[1]{\textcolor{black}{#1}}

\title{Gap Preserving Distillation by Building Bidirectional Mappings with A Dynamic Teacher}

\author{Yong Guo$^1$, Shulian Zhang$^2$, Haolin Pan$^2$, Jing Liu$^3$, Yulun Zhang$^4$, Jian Chen$^2$ \\
$^1$Max Planck Institute for Informatics\\
$^2$South China University of Technology\\
$^3$Monash University\\
$^4$Shanghai Jiao Tong University\\
}

\begin{document}

\maketitle

\begin{abstract}
Knowledge distillation aims to transfer knowledge from a large teacher model to a compact student counterpart, often coming with a significant performance gap between them. We find that a too-large performance gap can hamper the training process, which is also verified in recent studies.
To address this, we propose a \textbf{Gap Preserving Distillation (GPD)} method that trains an additional dynamic teacher model from scratch along with training the student to bridge this gap. In this way, it becomes possible to maintain a reasonable performance gap between teacher and student during the whole distillation process. 
\yong{To further strengthen distillation from the dynamic teacher to the student, we develop a hard strategy by enforcing them to share parameters and encouraging parameter inheritance. Besides hard strategy, we also build the soft bidirectional mappings between them which are built on an \emph{\textbf{Inverse Reparameterization (IR)}} method and a \emph{\textbf{Channel-Branch Reparameterization (CBR)}} strategy.}
We highlight that our IR is able to initialize a larger dynamic teacher with an arbitrary expansion ratio, while preserving exactly the same accuracy as the given student model. \yong{In this way, it guarantees that the dynamic teacher and student start from the same point and avoid a too large gap in early stage of training.}
As for our CBR, with parameter-sharing, it directly extracts an effective student model from the well-learned dynamic teacher without any post-training, making our method highly flexible for model deployment.
In the experiments, GPD significantly outperforms existing distillation methods on top of both CNNs and transformers architectures, achieving up to 1.58\% accuracy improvement. 
Interestingly, GPD also generalizes well to the scenarios without a pre-trained teacher, including training from scratch and fine-tuning, yielding a large improvement of 1.80\% and 0.89\% on ResNet18, respectively.

\end{abstract}

\section{Introduction} \label{sec:intro}
Deep neural networks have achieved remarkable success across various domains~\citep{gu2024mamba, DBLP:journals/corr/abs-2305-17888,DBLP:journals/corr/abs-2302-13971,Wang_2023_CVPR,Zhang_2023_CVPR}. However, the high accuracy of these models often comes at the cost of large model sizes. 
Recent studies have begun to explore methods to reduce model complexity.
In parallel to model compression techniques~\citep{pmlr-v202-liu23w,wei2022outlier,10094002, DBLP:conf/icml/XiaoLSWDH23}, knowledge distillation (KD)~\citep{DBLP:journals/corr/HintonVD15} offers a solution by transferring knowledge from complex, high-capacity models to simpler, more lightweight models, thereby achieving effective model compression.

The standard practice of KD is to train a smaller student model to mimic the behavior or predictions of a teacher model~\citep{
Sun_2024_CVPR, Jin_2023_CVPR, DBLP:conf/cvpr/ZhaoCSQL22,
Li_Li_Yang_Zhao_Song_Luo_Li_Yang_2023, 
DBLP:conf/cvpr/ZhaoCSQL22, DBLP:journals/tnn/PassalisTT21, DBLP:journals/corr/abs-1910-10699, DBLP:conf/iclr/ZagoruykoK17, DBLP:conf/iccv/HeoKYPK019, DBLP:conf/cvpr/Chen0ZJ21, DBLP:conf/iclr/0038MBT21, Peng_2019_ICCV}. 
\yong{However, existing methods often exploit a fixed pre-trained teacher model but it does not always guide the training of student in the most effective way.}
To be specific, it has been shown that it is often non-trivial for the student to obtain promising knowledge/improvements from the teacher when there is a very large gap between the teacher and student models, especially in the early training stage. In contrast, a weaker teacher, together with a smaller performance gap from the student, has been shown to be a better choice~\citep{Son_2021_ICCV, yang2019training, mirzadeh2020improved, DBLP:journals/corr/abs-2002-09168, NEURIPS2022_03e0712b}. \shulian{In fact, it can be theoretically proved by \citep{NEURIPS2022_03e0712b} that weak models ``have higher mutual information regarding the input'' compared to stronger teacher models, which can enhance knowledge distillation.}
\yong{Interestingly, this phenomenon can be intuitively understood, just like it is hard for a child to learn advanced mathematics directly from a senior professor. Instead, it would be better to teach him/her to count or learn elementary mathematics first. In other words, for distillation, a student model should learn from a teacher that has an appropriate knowledge/performance gap over it.}  
\yong{Nevertheless, even with a relatively weak teacher that has a small performance gap, a fixed model that is commonly used would eventually become useless since the student will gradually improve and surpass the teacher at some time. Thus, how to adaptively control the performance gap within a reasonable range throughout the whole distillation process is a critical problem.}

To address this, \yong{we seek to explicitly control the performance gap to boost knowledge transfer. However, it is non-trivial to determine what performance gap should be regarded as ``too-large''. Although the gap can be measured by some criteria, \emph{e.g.}, accuracy for classification, the threshold to distinguish a too-large gap is still ambiguous. To solve this, we cast the problem of determining a threshold into measuring the gap in terms of model size between student and teacher.}
We propose to introduce a learnable dynamic teacher (DT) model besides the pre-trained static teacher model, training it from scratch together with the student. Since DT is a larger model and often converges faster than the student, we are able to keep a promising performance gap between DT and student during training.
In addition, we hope to build a stronger connection between DT and student to transfer knowledge in a more explicit way, in order to further enhance the performance. To achieve this, we \yong{develop a hard strategy for distillation that} enforces DT and the student to share their parameters and encourage parameter inheritance. In addition to hard strategy, we also build a soft bidirectional mapping between them via a novel reparameterization method.

In this paper, we make the following key contributions: 
\textbf{1)} We propose a {\textbf{ Gap Preserving Distillation (GPD)}} method that enhances distillation performance by introducing an additional dynamic teacher (DT). We simultaneously train DT and the student to maintain a reasonable accuracy gap between them during the whole training process. \yong{In this way, it becomes possible to build a dynamic performance gap \emph{at every iteration} between the student and teacher, which is essentially different from existing work.}
\yong{\textbf{2)} We develop a {\textbf{hard strategy}} for distillation where the student and the teacher share the same set of parameters. The key idea is that, since the teacher is often easier to obtain high accuracy, it may also be easier for the student to get promising improvement if it directly inherits well-learned parameters from the teacher, as empirically shown in Table~\ref{tab:ablation_components}.}
\textbf{3)} We explicitly enhance knowledge transfer by building \emph{bidirectional mappings} between DT and the student via {\textbf{Inverse Reparameterization (IR)}} and {\textbf{Channel-Branch Reparameterization (CBR)}}. 
IR constructs the dynamic teacher model by expanding the student model with an arbitrary expansion ratio along both the channel and branch dimensions, while preserving the same accuracy as the student model. This guarantees that both DT and the student can start from the same initial point and thus avoid a too-large performance gap in early training stage. \yong{Interestingly, IR is not designed just for distillation, but a general initialization method for building any-size pre-trained models, yielding additional contributions to the community beyond distillation.}
On the other hand, our CBR seeks to extract an effective student model from the shared parameters with DT without any post-training. Unlike existing reparameterization approaches, CBR does not equivalently transform a model into a more compact one, 
but extracts an effective student out of DT.
\textbf{4)} 
In experiments, 
\yong{GPD consistently outperforms existing distillation methods on top of both CNN and transformer architectures. We emphasize
that GPD is very flexible in that it also generalizes well to other training settings, including both training from scratch and fine-tuning, which is rarely reported by other methods.}

\section{Related Work}

\textbf{Knowledge distillation.}
Knowledge distillation (KD)~\citep{DBLP:journals/corr/HintonVD15} transfers knowledge from a teacher to a smaller student model. Methods improve this by focusing on logits or intermediate features~\citep{Sun_2024_CVPR, Jin_2023_CVPR, DBLP:conf/cvpr/ZhaoCSQL22, Li_Li_Yang_Zhao_Song_Luo_Li_Yang_2023, DBLP:journals/tnn/PassalisTT21, DBLP:journals/corr/abs-1910-10699, DBLP:conf/iclr/ZagoruykoK17, DBLP:conf/iccv/HeoKYPK019, DBLP:conf/cvpr/Chen0ZJ21, heo2019knowledge, kim2018paraphrasing}.
Standard methods prioritize fully converged teachers with high performance, yet the performance gap can hinder knowledge transfer~\citep{NEURIPS2022_03e0712b, DBLP:journals/corr/abs-2002-09168,cho2019efficacy, DBLP:journals/corr/abs-1909-11723}.
Strategies to address this include using intermediate-stage teachers~\citep{cho2019efficacy,DBLP:journals/tcyb/ZhaoSDCD22}, pre-training student-friendly teacher model~\citep{DBLP:conf/aaai/YangXQY19, DBLP:conf/nips/ParkCJKH21,dong2024toward}, \shulian{introducing intermediate-sized assistant teachers~\citep{mirzadeh2020improved, Son_2021_ICCV}} 
or introducing auxiliary networks~\citep{DBLP:journals/ijon/GaoWW21}. These methods often rely on specially designed and pre-trained intermediate models.
\shulian{Feature-based methods like DTSKD~\citep{LI2024110422} and DiffKD~\citep{NEURIPS2023_cdddf13f} focus on bridging semantic gaps or denoising features. SCKD~\citep{Zhu_2021_ICCV} optimizes transfer using gradient similarity. Recent works refine soft labels~\citep{YUAN2024111915,rao2023parameter} or student's output entropy~\citep{ZHU2024110545} to enhance knowledge transfer.}
In contrast, our GPD constructs a trainable dynamic teacher based on the student model, maintaining an appropriate accuracy gap throughout distillation for effective knowledge transfer.
\\
\textbf{Reparameterization.}
Structural reparameterization~\citep{Ding_2021_CVPR,DBLP:conf/cvpr/Ding0MHD021} has gained attention in tasks such as compact model design~\citep{DBLP:journals/corr/abs-2010-11929}, architecture search~\citep{DBLP:journals/corr/abs-1912-04749,10219892}, and pruning~\citep{DBLP:journals/corr/abs-2007-03260}. RepVGG~\citep{DBLP:conf/cvpr/Ding0MHD021} transforms training-time structures into equivalent, simpler inference structures. Other methods like DiracNet~\citep{DBLP:journals/corr/ZagoruykoK17}, ACB\citep{Ding_2019_ICCV}, DO-Conv\citep{9779456}, and ExpandNet~\citep{NEURIPS2020_0e1ebad6} also achieve structural reparameterization. OREPA~\citep{DBLP:journals/corr/abs-2204-00826} reduces training costs by online reparameterization. 
In contrast, our CBR enables the extraction of an effective student model from the dynamic teacher, enhancing knowledge transfer to the compact student model.
\\
\textbf{Model expansion.}
Net2Net~\citep{DBLP:journals/corr/ChenGS15} pioneered functional-preserving model expansion. bert2BERT~\citep{chen2021bert2bert} applied this to Transformers, with other works focusing on depth growth~\citep{dong2020towards, DBLP:conf/iclr/ChangMHTB18, DBLP:journals/corr/abs-2011-13635, gong2019efficient}. 
Staged Training~\citep{shen2022staged} and LEMON~\citep{DBLP:journals/corr/abs-2310-07999} further expanded both width and depth. \shulian{ControlNet~\citep{zhang2023adding} duplicates the model and adds zero convolution layers to maintain equivalence, freezing the original parameters for fine-tuning. ControlNet~\citep{zhang2023adding} duplicates the model, adds zero convolution layers, and freezes original parameters for fine-tuning. In contrast, our GPD expands the model and enables adaptive switching between compact and expanded models during training, maintaining an appropriate performance gap.}


\begin{figure}[t]
  \centering
  \includegraphics[width=1\linewidth]{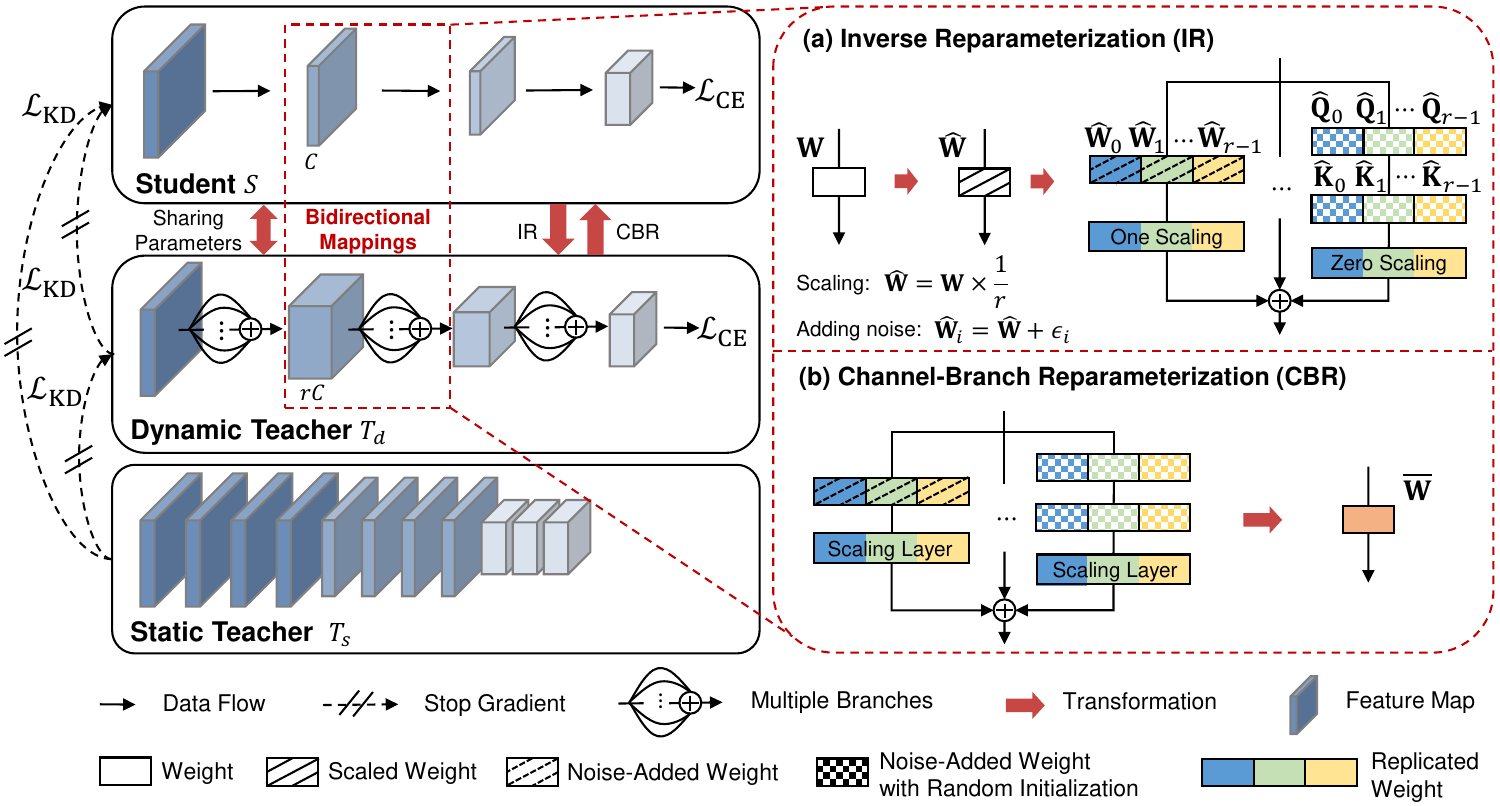}
  \caption{Overview of the proposed Gap Preserving Distillation (GPD) method. Besides the static teacher, we introduce an additional dynamic teacher and train it from scratch along with the student. The student model shares parameters with the dynamic teacher via \textbf{\emph{Inverse Reparameterization (IR)}} and \textbf{\emph{Channel-Branch Reparameterization (CBR)}}. 
  (a) The dynamic teacher is constructed through IR (top right) from the student model. For any layer, we replicate the weights along the channel dimension to build a wider layer while introducing additional branches to construct a multi-branch architecture. In order to maintain the same accuracy as the student, we only activate the first branch that contains the original student weights and zero out all the other extra branches, \emph{i.e.}, one-scaling and zero-scaling.  
  (b) We extract a promising student from the dynamic teacher via CBR. The expanded multi-branch architecture can be merged into the student's single-branch architecture using a similar way proposed by OREPA~\citep{DBLP:journals/corr/abs-2204-00826}. After that, given an expansion ratio $r$, we directly extract the first $1/r$ parameters multiplied by a scaling factor (see details in Section \ref{sec:channel_level_reparam}). 
  }
  \label{fig:overview}
  \vspace{-3 pt}
\end{figure}

\section{Gap Preserving Distillation} \label{sec:method}
\yong{In this work, we develop a \textbf{Gap Preserving Distillation (GPD)} method that enhances knowledge distillation by preserving an appropriate performance gap between the teacher and student throughout the whole distillation process.}
Instead of directly defining how large the gap should be, we propose to learn an additional dynamic teacher (DT) model as a proxy to maintain this gap, which is more flexible and controllable. 
Unlike existing methods, we build bidirectional mappings between DT and student to strengthen their connections. These two mappings can be achieved by the Inverse Reparameterization (IR) method and the Channel-Branch Reparameterization (CRB) method, respectively.
The overview of our method is shown in Figure~\ref{fig:overview} and Algorithm~\ref{alg: gpd}.

\subsection{Distillation with Dynamic Teacher}\label{sec:overview}
Popular KD methods often exploit a static teacher \(T_s\) to guide the training of the student $S$~\citep{Sun_2024_CVPR, Jin_2023_CVPR, DBLP:conf/cvpr/ZhaoCSQL22,
Li_Li_Yang_Zhao_Song_Luo_Li_Yang_2023, 
DBLP:conf/cvpr/ZhaoCSQL22, furlanello2018born, DBLP:conf/cvpr/ZhaoCSQL22, DBLP:journals/corr/RomeroBKCGB14, DBLP:journals/tnn/PassalisTT21, DBLP:journals/corr/abs-1910-10699, DBLP:conf/iclr/ZagoruykoK17, DBLP:conf/iccv/HeoKYPK019, DBLP:conf/cvpr/Chen0ZJ21, heo2019knowledge, kim2018paraphrasing}. Unlike them, our Gap Preserving Distillation (GPD) introduces a learnable dynamic teacher (DT) model \(T_d\), as shown in Figure~\ref{fig:overview}. Moreover, we seek to boost knowledge transfer via not only optimizing the distillation objectives, but also inheriting parameters from a better teacher model. To achieve this, we exploit parameter-sharing techniques~\citep{10102479, xie2021weight, zhang2022minivit,Ma_2022_CVPR, DBLP:conf/mm/Zhou0CDG022} and enforce \(T_d\) and $S$ to share the same set of parameters, which turns out to be particularly effective (see effectiveness in Section~\ref{sec:ablation}).

As for training, besides the standard objective of KD methods, we introduce an additional loss related to DT $\mathcal{L}_{\text{GPD}}$.
Let $\mathcal{L}_{\text{CE}}$ be the cross-entropy loss, and $S(x)$ denotes the student model's prediction based on input $x$. We use $\psi(\cdot)$ to represent a function that extracts desired knowledge from models, which could be either logits or features. $\mathcal{L}_{\text{KD}}$ measures the discrepancy between the knowledge of two models, with $\lambda$ controlling the distillation loss's importance. 
Given an image-label pair ${(x, y)}$, the objective function of GPD with a dynamic teacher is formulated as follows:
\begin{equation}
\begin{split}
\mathcal{L}_{\text{total}}= \underbrace{
\mathcal{L}_{\text{CE}}(S(x), y) + \lambda \mathcal{L}_{\text{KD}}(\psi(S(x)), \psi(T_s(x)))
}_{\text{standard objective of KD methods}} 
+ \mathcal{L}_{\text{GPD}}.
\end{split}
\end{equation}
\shulian{Where we simply set $\lambda=1$ in all experiments thus no additional hyperparameters are introduced.}
As for $\mathcal{L}_{\text{GPD}}$, in Figure~\ref{fig:overview}, we seek to use $T_d$ to guide the training of $S$ and thus introduce a KD loss $\mathcal{L}_{\text{KD}}(\psi(S(x)), \psi(T_d(x)))$. Regarding the training of $T_d$, we minimize both the cross-entropy loss $\mathcal{L}_{\text{CE}}$ and a KD loss between $T_d$ and $T_s$ via $\mathcal{L}_{\text{KD}}(\psi(T_d(x)), \psi(T_s(x)))$. Thus, $\mathcal{L}_{\text{GPD}}$ becomes
\begin{equation}
\label{eq: L_GPD}
\begin{split}
\mathcal{L}_{\text{GPD}}=\mathcal{L}_{\text{CE}}({T_d}(x), y) + \mathcal{L}_{\text{KD}}(\psi(S(x)), \psi(T_d(x))) + \mathcal{L}_{\text{KD}}(\psi(T_d(x)), \psi(T_s(x))).
\end{split}
\end{equation}
To avoid learning knowledge from a weaker model, we consider single-way knowledge transfer and apply stop-gradient on all the KD losses.

\begin{algorithm}[t]
    \caption{Training process of Gap Preserving Distillation (GPD).}
    \label{alg: gpd}
    \begin{algorithmic}[1]
        \Require Student $S$, static teacher $T_s$, epochs $N$, step size $\eta$, model parameters $\mathbf{W}$, weight of standard knowledge distillation loss $\lambda$, knowledge function $\psi(\cdot)$, training data $(x, y)$
        \State Obtain dynamic teacher $T_d$:
        $\mathbf{W}_d \leftarrow \text{IR}(\mathbf{W})$  // Inverse Reparameterization
        \For{$i=1$ to $N$}
            \State Forward propagation using the dynamic teacher via $\hat{y}_d = T_d(x)$;
            \State Compute gradients for dynamic teacher: \\
            $\quad\quad\quad \mathbf{G}_d \leftarrow \nabla_{\mathbf{W}_d}\mathcal{L}_{\mathrm{CE}}(\hat{y}_d, y) + \nabla_{\mathbf{W}_d}\mathcal{L}_{\mathrm{KD}}(\psi(\hat{y}_d),\psi(T_s(x)))$;
            \State Obtain student $S$ from dynamic teacher $T_d$: \\
            $\quad\quad\quad \mathbf{W}_s \leftarrow \text{CBR}(\mathbf{W}_d)$  // Channel-Branch Reparameterization
            \State Forward propagation using the student via $\hat{y}_s=S(x)$;
            \State Compute gradients for student: \\
            $\quad\quad\quad \mathbf{G}_s \leftarrow \nabla_{\mathbf{W}_s}\mathcal{L}_{\mathrm{CE}}(\hat{y}_s, y) + \lambda \nabla_{\mathbf{W}_s}\mathcal{L}_{\mathrm{KD}}(\psi(\hat{y}_s),\psi(T_s(x)))$\\
            $\quad\quad\quad\quad\quad\quad\quad + \nabla_{\mathbf{W}_s}\mathcal{L}_{\mathrm{KD}}(\psi(\hat{y}_s),\psi(T_d(x)))$;
            \State Update parameters $\mathbf{W}_d$ sharing between student $S$ and dynamic teacher $T_d$:\\
            $\quad\quad\quad \mathbf{W}_d \leftarrow \mathbf{W}_d - \eta(\mathbf{G}_d + \mathbf{G}_s)$;
        \EndFor
    \end{algorithmic}
\end{algorithm}

\begin{figure}[t]
\centering
\includegraphics[width=1\linewidth]{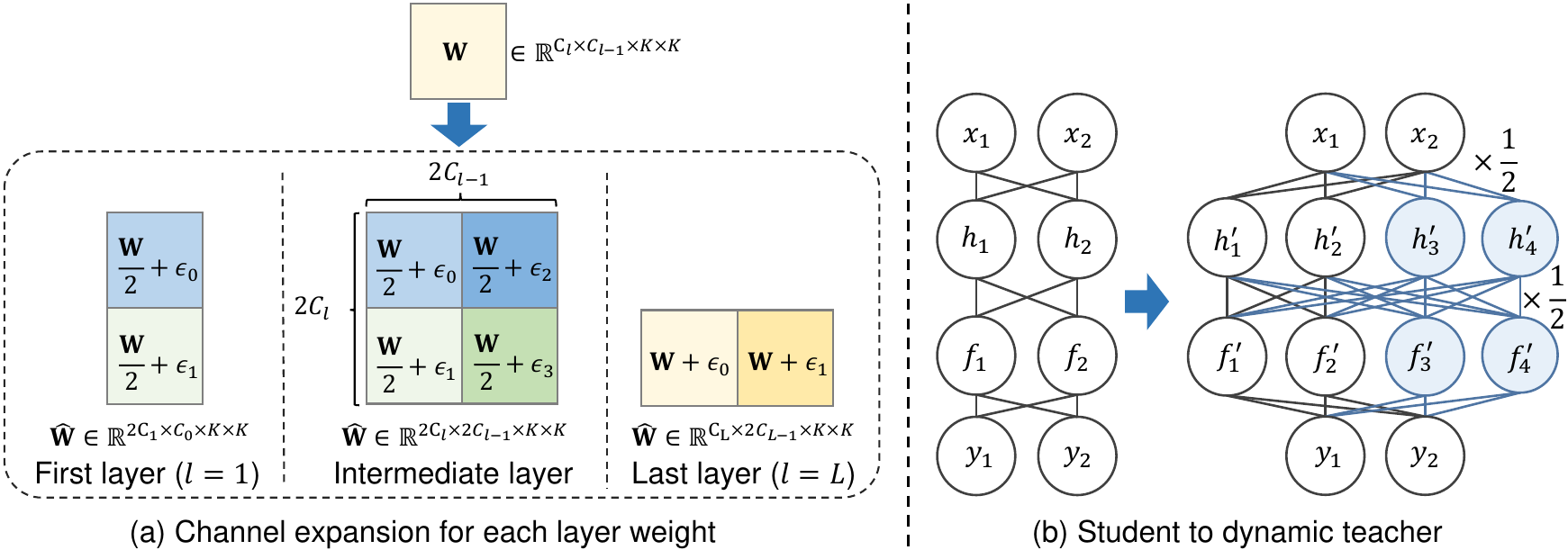}
\vspace{-15 pt}
\caption{Illustration of channel-level inverse reparameterization with an expansion ratio of 2. (a) For the first layer, weights are scaled by 2 and replicated along the output channel dimension, expanding from $C_1 \times C_0$ to $2C_1 \times C_0$. For intermediate layers, weights are scaled by 2, then replicated along both input and output dimensions, expanding from $C_l \times C_{l-1}$ to $2C_l \times 2C_{l-1}$. For the last layer, weights are replicated along the input dimension, expanding from $C_L \times C_{L-1}$ to $C_L \times 2C_{L-1}$. (b) Inverse re-parameterizing the student model (left) to construct the dynamic teacher model (right) by expanding channels from 2 to 4 following the procedures exemplified in (a), while preserving the initial input-output mapping.}
\vspace{-15 pt}
\label{fig:channel_inverse_reparam}
\end{figure}

\subsection{Build Dynamic Teacher via Inverse Reparameterization} \label{sec:inverse_reparam}
In order to build a suitable dynamic teacher, we develop an Inverse Reparameterization (IR) method to build a larger model from the student with any expansion ratio $r$, see Figure~\ref{fig:overview} (top right). To achieve this, we expand the model along both the channel and branch dimensions. One key characteristic is that the expanded dynamic teacher shares exactly the same accuracy as the student. 
\shulian{Notably, IR can serve as a general initialization method for building any-size pre-trained models, yielding additional contributions to the community beyond distillation.}

\subsubsection{Channel-Level Inverse Reparameterization}
\label{sec:channel_level_inverse_reparameterization}
In the channel level, we seek to expand the number of channels in each layer by an expansion ratio $r$ without sacrificing accuracy. To this end, we propose a channel-level inverse reparameterization strategy, as shown in Figure~\ref{fig:channel_inverse_reparam}. The key idea is to replicate the weights of the student and introduce a scale factor to compensate for the increased number of channels.

Considering a student model with $L$ convolutional layers, let $\mathbf{W}^l \in \mathbb{R}^{C_l \times C_{l-1} \times K \times K}$ be the weight of the $l$-th layer. Here, $C_l$ and $C_{l-1}$ represent the number of output and input channels, and $K \times K$ denotes the kernel size.
Let $\mathbf{\widehat{W}}^l \in \mathbb{R}^{rC_l \times rC_{l-1} \times K \times K}$ be the expanded weight matrix for layer $l$ of the dynamic teacher, given the expansion ratio $r$.
To better illustrate our method, we divide all layers into three groups, including the first layer, intermediate layers, and the last layer.
In Figure~\ref{fig:channel_inverse_reparam}, we take the expansion ratio $r=2$ for example to illustrate our method.
For the first layer ($l=1$), 
we replicate $\mathbf{W}^1$ by $r$ times and scale them by $1/r$. In this way, given the same input $x$, the output would be $1/r$ of the original one. Nevertheless, since the number of output channels has been extended from $C_1$ to $rC_1$, summing up all the channels would obtain the same value as the original output. For convenience, we use $\mathbf{\widehat{W}}^1 \in \mathbb{R}^{rC_1 \times C_0 \times K \times K}$ to denote the weight scaled by $1/r$. For the last layer ($l=L$), the original $\mathbf{W}^L$ is replicated $r$ times along the input dimension, giving $\mathbf{\widehat{W}}^L \in \mathbb{R}^{C_L \times rC_{L-1} \times K \times K}$.
As for the intermediate layers,
we scale $\mathbf{W}^l$ by $1/r$ and replicate the scaled weights $r$ times along both output and input channel dimensions, yielding $rC_l$ output channels and $rC_{l-1}$ input channels. To avoid the trivial solution caused by symmetrical/identical replications during training, we introduce a small noise $\epsilon$ into each replication.
The theoretical proof of the equivalence of channel-level inverse reparameterization is provided in Appendix~\ref{app:proof}.

\subsubsection{Branch-Level Inverse Reparameterization}
Besides channel-level expansion, we also seek to expand the student model along the branch dimension.
The branch-level inverse reparameterization aims to expand a single convolutional layer into a multi-branch structure with increased capacity, while preserving the identical input-output mapping. The key idea is to introduce additional branches with extra convolutions but zero-out them to keep the output of multi-branch architecture the same as the single-branch counterpart.

Specifically, considering a convolutional layer represented by the kernel $\mathbf{W} \in \mathbb{R}^{C_l \times C_{l-1} \times K \times K}$, where $\mathbf{X} \in \mathbb{R}^{C_{l-1} \times H \times W}$ and $\mathbf{Y} \in \mathbb{R}^{C_l \times H' \times W'}$ are the input and output tensors, respectively. Thus, the convolution operation can be represented by $\mathbf{Y} = \mathbf{W}\mathbf{X}$. As shown in Figure~\ref{fig:overview} (top right), we expand this single convolution into a multi-branch topology with $M$ branches.
The first branch consists of a single convolutional layer, whose weights are initialized with the original convolution weights $\mathbf{W}$, followed by a learnable linear scaling layer $\mathbf{S}_1 \in \mathbb{R}^{C_l}$ initialized as a vector of ones.
For the remaining $M-1$ branches, each branch $m$ comprises a stack of convolutional layers, with their weights randomly initialized. All these branches are also followed by a learnable linear scaling layer $\mathbf{S}_m \in \mathbb{R}^{C_l}$, initialized with a vector of zeros. 
In this way, the output of the multi-branch block is equivalent to the original single-branch convolution.
Let $f_m(\mathbf{X})$ denote the computation of the $m$-th branch. The expanded model along the branch dimension can be formulated by
\begin{equation}
\mathbf{Y} = \sum_{m=1}^{M}\mathbf{S}_mf_m(\mathbf{X})=\mathbf{S}_1f_1(\mathbf{X}) = \mathbf{WX}.
\end{equation}

\subsection{Parameter Sharing with Channel-Branch Reparameterization}\label{sec:param_sharing}

\begin{figure}[t]
  \centering
  \vspace{-10 pt}
  \includegraphics[width=1\linewidth]{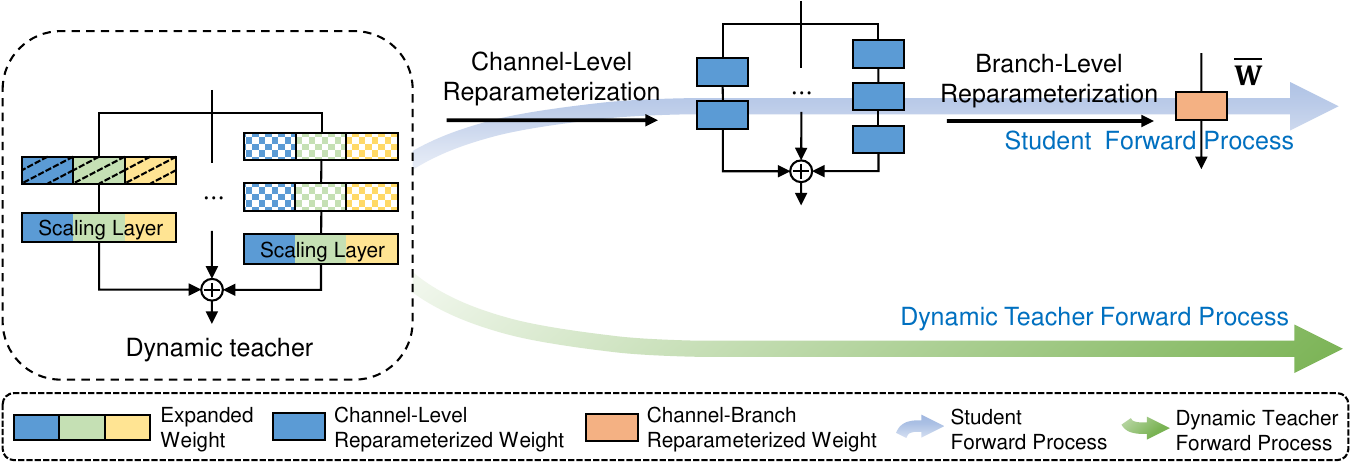}
  \vspace{-15 pt}
  \caption{Illustration of the forward process for the student and dynamic teacher models. The dynamic teacher performs a direct forward pass, utilizing its increased capacity. The student model shares all parameters from the dynamic teacher and undergoes a two-step reparameterization process. First, channel-level reparameterization adjusts the expanded channels to match the original channel dimensions of the student model. Second, branch-level reparameterization merges the expanded multi-branch units into a single branch structure, thereby restoring the original topology of the student model while inheriting knowledge from the dynamic teacher.}
  \label{fig:forward_process}
  \vspace{-10 pt}
\end{figure}

To further enhance knowledge transfer between the student $S$ and the dynamic teacher $T_d$, we propose a hard strategy for distillation that forces them to share parameters. In this way, the student is able to directly inherit well-trained parameters from a larger model, \emph{i.e.}, dynamic teacher $T_d$. Based on the shared parameters, $S(x)$ and $T_d(x)$ have different predictions due to different forward propagation methods, as shown in Figure~\ref{fig:forward_process}. Specifically, $T_d$ directly performs the forward pass, while the student $S$ conducts online reparameterization in both channel-level and branch-level.

\subsubsection{Channel-Level Reparameterization}
\label{sec:channel_level_reparam}

As for channel-level reparameterization, given an expansion ratio $r$, we directly take the first $1/r$ kernels and scale them by $r$. Interestingly, this mapping is exactly the inverse transformation of how to conduct channel-level inverse parameterization (previously discussed in Section~\ref{sec:channel_level_inverse_reparameterization}). Formally, 
let $\mathbf{W}^l_m \in \mathbb{R}^{rC^l_m \times rC^{l-1}_m \times K \times K}$ denote the weight of the $l$-th convolutional layer in the $m$-th branch of an expanded layer, and $rC^l_m$ and $rC^{l-1}_m$ are the numbers of output and input channels, respectively.
\shulian{During training, our channel-level reparameterization strategy explicitly extracts a subset of the dynamic teacher's expanded parameters to construct the student model's weights, enabling us to obtain a promising student model after training.}  Specifically, we extract a channel-wise slice from $\mathbf{W}^l_m$ and apply a scaling operation:
\begin{equation}
\mathbf{\bar{W}}^l_m = r \mathbf{W}^l_m[:C^l_m, :C^{l-1}_m, :, :],
\end{equation}
where $\mathbf{\bar{W}}^{l}_m \in \mathbb{R}^{C^l_m \times C^{l-1}_m \times K \times K}$ denotes the reparameterized weight for the corresponding layer in the student model. The slicing operation $[:C^l_m, :C^{l-1}_m, :, :]$ extracts the first $C^l_m$ output channels and $C^{l-1}_m$ input channels from $\mathbf{W}^l_m$. The scaling ratio $r$ ensures the extracted parameters are appropriately scaled, aligning with the IR process applied during the construction of the dynamic teacher.
For the first and last layers, the channel extraction is performed only over the output or input channel dimensions, respectively. Specifically, for the first layer, $\bar{\mathbf{W}}^1_m = r \mathbf{W}^1_m[:C^1_m, :, :, :]$, while for the last layer, $\bar{\mathbf{W}}^L_m = \mathbf{W}^L_m[:, :C^{L-1}_m, :, :]$, as the last layer's weights are not scaled during IR process.
Moreover, we found that parameter sharing with Batch Normalization~\citep{ioffe2015batch} layers requires special treatment, as detailed in Appendix~\ref{app:bn}.

\subsubsection{Branch-Level Reparameterization}
After channel-level reparameterization, we merge the expanded multi-branch units of the dynamic teacher model into the student's single-branch structure via branch-level reparameterization. Following \citet{DBLP:journals/corr/abs-2204-00826}, for the $m$-th branch that contains $L_m$ convolutional layers ${\mathbf{W}^1_m, \mathbf{W}^2_m, ..., \mathbf{W}^{L_m}_m}$, we first obtain ${\mathbf{\bar{W}}^1_m, \mathbf{\bar{W}}^2_m, ..., \mathbf{\bar{W}}^{L_m}_m}$ through channel-level reparameterization. 
\shulian{These are then merged into a single weight $\mathbf{\overline{W}}_m$ via the standard reparameterization, which is mathematically equivalent, as proven in~\citep{DBLP:conf/cvpr/Ding0MHD021}, by conducting sequential convolution operations:} $\mathbf{\overline{W}}_m = \mathbf{\bar{W}}^1_m \mathbf{\bar{W}}^2_m ...\mathbf{\bar{W}}^{L_m}_m$. 
Performing this for all $M$ branches yields ${\mathbf{\overline{W}}_1, \mathbf{\overline{W}}_2, ..., \mathbf{\overline{W}}_M}$. After that, we sum up all of them to get the final new weight $\mathbf{\overline{W}} = \sum_{m=1}^M \mathbf{\overline{W}}_m$ for this layer in the student model. 
Through the branch reparameterization strategy, the student model effectively inherits well-trained parameters from the dynamic teacher model.

\begin{table}[t]
    \vspace{-10 pt}
    \caption{Comparison of the performance of various distillation methods across different architectures. ``-'' denotes the result that is not reported. A $\rightarrow$ B indicates a teacher model A distilling knowledge to a student model B. GPD consistently enhances the performance of standard distillation methods across diverse architectures.}
    \label{tab:sota_comparison}
    \center
    \resizebox{0.97\linewidth}{!}
    {
    \begin{tabular}{c|c|c|c}
        \toprule
        \multirow{2}{*}{Model} & \multicolumn{3}{c}{Teacher $\rightarrow$ Student} \\
        \cmidrule{2-4}
        
        & \multicolumn{1}{c|}{ResNet34 $\rightarrow$ ResNet18} & \multicolumn{1}{c|}{ResNet50 $\rightarrow$ MobileNet} & \multicolumn{1}{c}{RVT-S $\rightarrow$ RVT-Ti} \\

        \midrule
        Teacher & 73.31   & 76.16   & 81.69   \\
        Student & 69.75    & 68.87   & 78.45 \\ 
        \midrule
        KD~\citep{DBLP:journals/corr/HintonVD15} & 70.66  & 68.58       &  -     \\
        AT~\citep{DBLP:conf/iclr/ZagoruykoK17} & 70.69    & 69.56    &   -   \\
        OFD~\citep{DBLP:conf/iccv/HeoKYPK019} & 70.81  &71.25      &   -    \\
        CRD~\citep{DBLP:journals/corr/abs-1910-10699} & 71.17  & 71.37  & -   \\
        RKD~\citep{Park_2019_CVPR} & 70.40  & 68.5  & -   \\
        WSLD~\citep{zhou2021rethinking} & 72.04  & 71.52  & -   \\
        SRRL~\citep{DBLP:conf/iclr/0038MBT21} & 71.73 & 72.49 & -  \\
        SimKD~\citep{Chen_2022_CVPR} & 71.59 & 72.25 & -  \\
        DIST~\citep{huang2022knowledge} & 72.07 & 73.24 & -  \\
        NKD~\citep{Yang_2023_ICCV} & 71.96 & 72.58 & -  \\
        CAT-KD~\citep{Guo_2023_CVPR} & 71.26 & 72.24 & -  \\
        KD+CTKD~\citep{Li_Li_Yang_Zhao_Song_Luo_Li_Yang_2023} & 71.38 & 71.16 & -  \\
        MLKD~\citep{Jin_2023_CVPR} & 71.90 & 73.01 & -  \\
        KD+CTKD+LS~\citep{Sun_2024_CVPR} & 71.81 & 72.92 & -\\
        DKD+LSKD~\citep{Sun_2024_CVPR} & 71.88 & 72.85 & -  \\
        MLKD+LSKD~\citep{Sun_2024_CVPR} & 72.08 & 73.22 & -  \\
        CKD~\citep{DBLP:journals/corr/abs-2404-14109} & 72.24 & 72.97 & -  \\
        \midrule
        ReviewKD~\citep{DBLP:conf/cvpr/Chen0ZJ21}  & 71.61    & 72.56  &  78.92 \\
        {ReviewKD + GPD}   &   \textbf{72.50 (+0.89)}     & \textbf{73.21 (+0.65)} &  \textbf{80.01 (+1.09)}        \\
        \midrule
        DKD~\citep{DBLP:conf/cvpr/ZhaoCSQL22}   & 71.70    & 72.05   & 79.12  \\
        {DKD + GPD}                   &    \textbf{72.71 (+1.01)}       &  \textbf{73.63 (+1.58)}   &    \textbf{80.14 (+1.02)}         \\
        \bottomrule
    \end{tabular}
    }
    \vspace{-10 pt}
\end{table}

\subsection{Advantages over existing methods} 
Our GPD has specific advantages and is essentially different from existing distillation methods~\citep{NEURIPS2022_03e0712b, mirzadeh2020improved, cho2019efficacy, DBLP:conf/aaai/YangXQY19, Son_2021_ICCV, DBLP:journals/tcyb/ZhaoSDCD22} in several aspects. \textbf{First}, they maintain different performance gaps. Existing methods~\citep{dong2024toward,mirzadeh2020improved, Son_2021_ICCV} utilize fixed-accuracy pre-trained teachers assistant models, which may hinder knowledge transfer as the gap still be too large, particularly in early training stages when student accuracy is low. GPD initializes the dynamic teacher with the same accuracy as the student and trains both simultaneously, maintaining a small, dynamic gap throughout the process.
\textbf{Second}, they construct the teacher assistants in different ways. Unlike methods~\citep{mirzadeh2020improved, Son_2021_ICCV} that construct and train multiple intermediate-size teacher assistants separately, which can be computationally expensive, GPD builds a single dynamic teacher. We propose a novel IR technique for model expansion that maintains the same initial accuracy as the student. This approach is both computationally efficient and ensures a controlled start point for the distillation process.
\textbf{Third}, they transfer knowledge in different ways. Existing methods~\citep{mirzadeh2020improved, Son_2021_ICCV} follow the standard distillation paradigm by incorporating a KD loss for transferring knowledge. Besides this, our GPD enforces parameter sharing between the student and dynamic teacher, allowing direct inheritance of parameters. This process is facilitated by our CBR, enabling a more direct and effective knowledge transfer.

\section{Experiments} \label{sec: exp}
We evaluate GPD on ImageNet~\citep{5206848} using popular classification models. We first compare GPD with state-of-the-art KD methods across CNNs and transformers in Sec.~\ref{sec:distillation_with_ST}. 
We then explore two scenarios without a static teacher, including training models from scratch (Sec.~\ref{sec:train_from_scratch_wo_ST}) and fine-tuning (Sec.~\ref{sec:fine-tuning}). 
For clarity, we use GPD* to denote our models in the settings without a pre-trained static teacher. \shulian{Detailed experimental settings are in Appendix~\ref{app:setting}.}

\subsection{Distillation with A Static Teacher}\label{sec:distillation_with_ST}

We closely follow the settings of \citet{DBLP:conf/cvpr/ZhaoCSQL22},  \citet{DBLP:conf/cvpr/Chen0ZJ21} and put details in Appendix.
Table~\ref{tab:sota_comparison} shows the consistent superiority of our GPD across diverse architecture when training from scratch with a static teacher. 
\shulian{Notably, GPD not only yields substantial accuracy improvements when combined with existing KD methods like ReviewKD and DKD, but also outperforms the latest state-of-the-art KD approaches.}
Specifically, in the ResNet34 $\rightarrow$ ResNet18 setting, GPD boosts ReviewKD from 71.61\% to 72.50\%, and improves DKD by 1.01\%, reaching 72.71\%. \shulian{These results outperform the most recent methods.}
In ResNet50 $\rightarrow$ MobileNet, GPD enhances ReviewKD by 0.65\% and DKD by 1.58\%. Similar gains are also observed in the transformer-based setting RVT-S $\rightarrow$ RVT-Ti.
These significant accuracy improvements across diverse architectures highlight the effectiveness of our proposed method. By introducing a dynamic teacher model to mitigate the substantial gap between the student and a powerful static teacher, our GPD enables the student to more effectively absorb knowledge from the teacher. 
\shulian{This leads to substantial performance improvements when combined with existing KD methods, surpassing even the most recent distillation techniques.}

\subsection{Train from Scratch} \label{sec:train_from_scratch_wo_ST}
In this experiment, we train ResNet18 and MobileNet for 100 epochs, and RVT-Ti for 300 epochs. Table~\ref{tab:train_from_scratch_wo_ST} illustrates the effectiveness of our GPD* method, which utilizes only the dynamic teacher for distillation, without a static teacher. Across various backbone architectures, GPD* consistently improves upon baseline models trained without knowledge distillation. For example, the ResNet-18 model achieves a significant accuracy boost from 70.07\% to 71.87\% with GPD*, indicating a notable 1.80\% improvement. Similarly, with GPD*, the MobileNet achieves a noteworthy improvement from 71.68\% to 73.07\%. 
These results highlight the versatility and effectiveness of our proposed method,
which not only enhances the performance of existing KD methods when a strong static teacher is available but also serves as an effective stand-alone training strategy in scenarios where pre-trained teacher models are unavailable. 

\begin{table}[h]
  \vspace{-20 pt}
  \caption{Performance comparison of training from scratch without a static teacher model. GPD* denotes our method using only the dynamic teacher for distillation. 
  As a standalone method, our GPD* consistently improves student model performance across diverse architectures.}
  \label{tab:train_from_scratch_wo_ST}
    \center
    \resizebox{0.58\linewidth}{!}
    {
    \begin{tabular}{c|c|c|c}
        \toprule
        Model & ResNet18 & MobileNet & RVT-Ti \\
    
        \midrule
        Baseline & 70.07 & 71.68 & 78.45 \\
        GPD* & \textbf{71.87 (+1.80)} & \textbf{73.07 (+1.39)} & \textbf{79.85 (+1.40)} \\
    \bottomrule
  \end{tabular}
  }
  \vspace{-5 pt}
\end{table}

\subsection{Model Fine-Tuning} \label{sec:fine-tuning}

In this experiment, we fine-tune the pre-trained models for 50 epochs and set the initial learning rate to 0.1$\times$ \wrt~its base/standard value.
As for our GPD*, the dynamic teacher is constructed via Inverse Reparameterization based on the student model itself. This process ensures that the dynamic teacher initially exhibits the same accuracy as the student.
In Table~\ref{tab:finetune},
compared to the accuracy of pre-trained model, GPD* consistently achieves performance improvements across different architectures. Moreover, GPD* outperforms the longer training approach by up to 0.89\% for ResNet18, and similar performance gains are observed for MobileNet and RVT-Ti models as well.

\begin{table}[h]
  \vspace{-15 pt}
  \caption{Performance comparison of fine-tuning. GPD* denotes distillation using only the dynamic teacher model without the static teacher. Our method consistently outperforms the fine-tuning baseline with longer training across various architectures.}
  \label{tab:finetune}
    \center
    \resizebox{0.8\linewidth}{!}
    {
    \begin{tabular}{c|c|c|c}
        \toprule
        Model & ResNet18 & MobileNet & RVT-Ti \\
        \midrule
        Pretrained Model & 69.75 & 68.87 & 78.45 \\
        \midrule
        Fine-Tuning with Longer Training & 70.23 & 69.01 & 78.61 \\
        GPD* & \textbf{71.12 (+0.89)} & \textbf{69.47 (+0.46)} & \textbf{78.84 (+0.23)} \\
        \bottomrule
  \end{tabular}
  }
  \vspace{-10 pt}
\end{table}

\section{Further Discussions} \label{sec:ablation}
\shulian{In this section, we present further ablation experiments to analyze the contributions of various components to performance gain. We also validate that too-large teacher models can hinder distillation. For a discussion on the computational cost of our GPD, please refer to Appendix~\ref{app:overhead}.}

\textbf{Gap preserving and parameter sharing.}
We investigate the contribution of the proposed accuracy gap preserving mechanism and the parameter sharing strategy in this part.
The ablation study in Table~\ref{tab:ablation_components} validates the critical role of these components in boosting performance. Individually, the gap preservation mechanism yields a 0.66\% accuracy gain, demonstrating its effectiveness in guiding the student model's learning trajectory. When combined with parameter sharing, the synergistic effect leads to a substantial 1.01\% improvement over the baseline, underscoring the significance of these key innovations in facilitating knowledge transfer within the GPD framework.

\textbf{Channel-branch reparameterization.}
Table~\ref{tab:ablation_reparam_level} provides insights into the impact of different inverse reparameterization strategies on the performance of our GPD. Both channel-level and branch-level reparameterization techniques individually contribute to performance improvements over the baseline DKD method, achieving accuracy gains of 0.86\% and 0.61\%, respectively. However, their combined application yields the highest performance boost, with a remarkable 1.01\% accuracy gain.
Utilizing both techniques enables the dynamic teacher to effectively guide the student model, maximizing knowledge transfer within the GPD.

\begin{table}[h]
    \vspace{-20pt}
    \centering
    \hspace*{-10pt}
    \label{tab:ablation}
    \begin{minipage}{0.47\linewidth}
            \caption{Ablation studies on preserving the accuracy gap and parameter sharing. We take DKD as the baseline method to distill knowledge from ResNet34 to ResNet18. Preserving the accuracy gap alone improves performance over the baseline, and combining it with parameter sharing yields further gains.}
        \begin{center}
        \resizebox{0.95\linewidth}{!}
        {
        \begin{tabular}{cc|c}
        \toprule
                    Preserve Gap   & Share Param     & Acc. (\%) \\
        \midrule
        \multicolumn{2}{c|}{Baseline}   & 71.70 \\
        \midrule
        \checkmark  &    &   72.36 (+0.66)  \\
        \checkmark  &\checkmark  &  \textbf{72.71 (+1.01)}   \\
        \bottomrule
        \end{tabular}
        }
        \end{center}
        \label{tab:ablation_components}
    \end{minipage}%
    ~~~~~~~
    \begin{minipage}{0.47\linewidth}
            \caption{Impact of inverse reparameterization level on performance. We take DKD as the baseline method to distill knowledge from ResNet34 to ResNet18. The combined channel-level and branch-level strategy achieves the highest accuracy.}
        \begin{center}
        \resizebox{0.95\linewidth}{!}
        {
        \begin{tabular}{cc|c}
            \toprule
            Channel-level & Branch-level  &  Acc. (\%)\\
            \midrule
                \multicolumn{2}{c|}{Baseline}   & 71.70 \\
                \midrule
                \checkmark    &               & 72.56 (+0.86) \\
                        & \checkmark    & 72.31 (+0.61) \\
                \checkmark    & \checkmark    & \textbf{72.71 (+1.01)}  \\
        
            \bottomrule
        \end{tabular}
        }
        \end{center}
        \label{tab:ablation_reparam_level}
    \end{minipage}
\end{table}

\begin{figure}[t]
    \vspace{-10pt}
    \centering
    \includegraphics[width=1\textwidth]{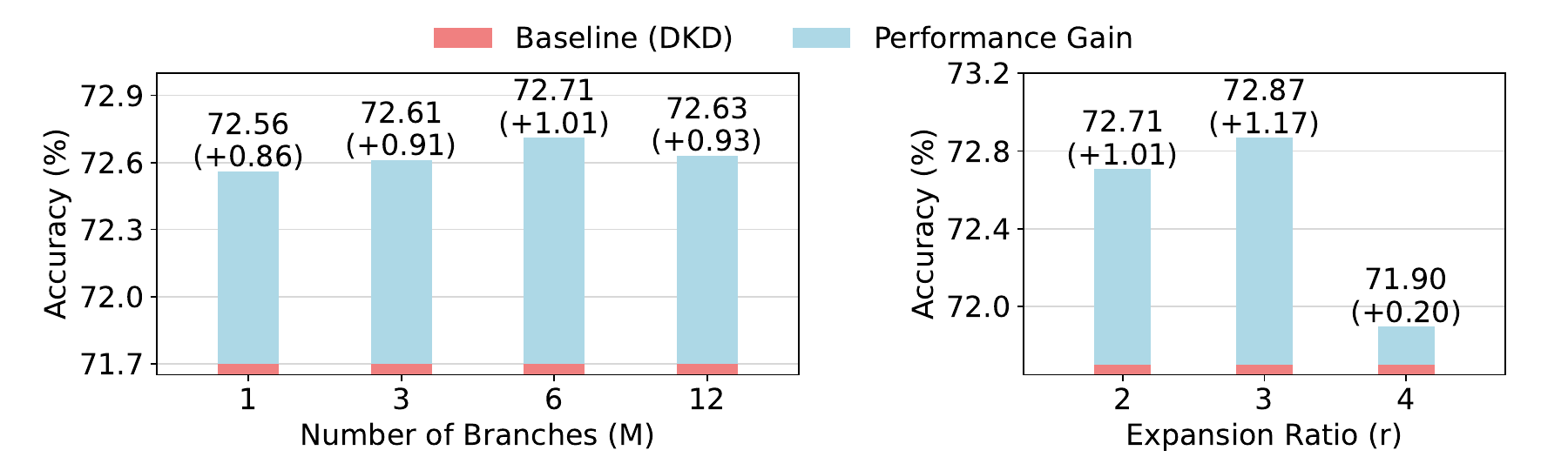} 
    \vspace{-20pt}
    \caption{ 
    \shulian{
    Impact of branch expansion number and channel expansion ratio on model accuracy.
    Performance gains are shown above the baseline (DKD). 
    Left: Increasing $M$ to 6 yields significant improvements, with diminishing returns beyond that. Right: Channel expansion ratios of 2 and 3 show substantial gains, while a ratio of 4 leads to degradation. 
    }}
    \vspace{-15pt}
    \label{fig:expansion_ratio}
\end{figure}

\shulian{\textbf{Branch and channel expansion ratio.}
Figure~\ref{fig:expansion_ratio} (left) shows that increasing the number of branches from 1 to 6 gradually improves accuracy, peaking at $M=6$. However, further increasing to 12 branches leads to reduced performance gains, indicating that too many branches may be hard to train and do not necessarily improve performance.
Figure~\ref{fig:expansion_ratio} (right) illustrates the impact of the channel expansion ratio $r$. Both $r=2$ and $r=3$ significantly outperform the DKD baseline, but improvement dramatically drops at $r=4$. The main reason is that the performance gap would be very large again when we consider a large dynamic teacher.
In practice, we recommend using $M=6$ and $r=2$ for a good balance between performance and efficiency.}

\shulian{\textbf{Teacher model size.}}
\shulian{We highlight that a larger/better teacher model does not necessarily improve the distillation performance, which is a common observation and has been theoretically proved~\citep{NEURIPS2022_03e0712b}. To verify this, we distill ResNet18 from two large teacher models, including ResNet101 and ViT-L~\citep{DBLP:conf/iclr/DosovitskiyB0WZ21}. From Table~\ref{tab:teacher_size}, ResNet101 outperforms ResNet34 by 4.06\% in accuracy but obtains similar distillation results based on DKD (71.70 vs. 71.74). As for a larger teacher ViT-L, it outperforms ResNet34 by 11.84\% but yields even worse results. More critically, no matter what kinds of teachers are used, GPD consistently improves the distillation performance by at least 1.01\%.}
\begin{table}[h]
    \vspace{-15 pt}
    \caption{
    \shulian{Comparison of distillation performance for ResNet18 with teacher models of varying sizes. Notably, larger teacher models, such as ViT-L, can hinder distillation performance, as observed with DKD. In contrast, our GPD consistently enhances results across all settings.
    }
    }
    \label{tab:teacher_size}
    \center
    \resizebox{0.9\linewidth}{!}
    {
    \begin{tabular}{c|c|c|c}
        \toprule
        \multirow{1}{*}{Method} & \multicolumn{1}{c|}{ResNet34 $\rightarrow$ ResNet18} & \multicolumn{1}{c|}{ResNet101 $\rightarrow$ ResNet18} & \multicolumn{1}{c}{ViT-L $\rightarrow$ ResNet18} \\

        \midrule
        Teacher & 73.31   & 77.37   & 85.15   \\
        Student & 69.75    & 69.75   & 69.75 \\ 
        \midrule
        DKD & 71.70 & 71.74 & 71.43 \\
        {DKD + GPD}  &    \textbf{72.71 (+1.01)}       &  \textbf{72.90 (+1.16)}   &    \textbf{72.71 (+1.28)}         \\
        \bottomrule
    \end{tabular}
    }
\end{table}\vspace{-15 pt}

\section{Conclusion} \label{sec:conclusion}
In this paper, we propose Gap Preserving Distillation (GPD), a novel approach to bridging the accuracy gap between teacher and student models for more effective knowledge transfer. Our key contribution is the introduction of a dynamic teacher model that preserves an appropriate accuracy lead over the student during training. We propose Inverse Reparameterization to losslessly expand the student model along the channel and branch dimensions, constructing the dynamic teacher with increased capacity. Furthermore, we devise a parameter sharing strategy based on Channel-Branch Reparameterization, enabling the student to inherit parameters from the expanded dynamic teacher. This reduces computational costs while allowing the student to benefit from the teacher's enriched knowledge representations. 
The improved efficiency and performance of compact models facilitated by our approach could enable the deployment of deep learning solutions in resource-constrained environments, thereby promoting wider accessibility to AI technologies.
Comprehensive experiments on the ImageNet dataset validate the effectiveness of GPD in boosting the performance of standard knowledge distillation methods across various backbone architectures. 


\bibliography{iclr2025_conference}
\bibliographystyle{iclr2025_conference}
\appendix
\section*{Appendix}
\section{Overview and Outline}
In this paper, we propose Gap Preserving Distillation (GPD) to bridge the performance gap between large teacher and compact student models. GPD trains a dynamic teacher alongside the student, maintaining a reasonable gap throughout. We utilize parameter sharing and establish mappings via \emph{Inverse Reparameterization (IR)} and \emph{Channel-Branch Reparameterization (CBR)}. The supplementary material provides detailed theoretical analyses and additional experimental information to support the main paper, organized as follows:
\begin{itemize}
    \item In Section~\ref{app:proof}, we provide a mathematical proof of channel-level inverse reparameterization, demonstrating the equivalence between the expanded model and the original model through detailed derivations.
    \item In Section~\ref{app:bn}, we discuss our strategy for parameter sharing in batch normalization, illustrating the importance of maintaining independent running statistics for student and dynamic teacher models to ensure stable training.
    \item In Section~\ref{app:setting}, we detail our experimental settings, covering various training scenarios including distillation with static teachers, training from scratch and fine-tuning. 
    \item In Section~\ref{app:overhead}, we evaluate the computational overhead introduced by our proposed methods, providing comparative analyses that highlight the efficiency of our approach.
\end{itemize}

\shulian{
\section{Proof of Channel-Level Inverse Reparameterization} \label{app:proof}
To mathematically prove the equivalence of channel-level inverse reparameterization, consider a model with three convolutions (see Figure 2(b)), where the parameters are denoted as $W^1$, $W^2$, and $W^3$. Using $\otimes$ to represent the convolutional operation, the computation of this model becomes}
{\small
\begin{equation}
  Y = W^3 \otimes W^2 \otimes W^1 \otimes X.
\end{equation}
}

\shulian{Following the expansion rules in Figure 2(a),  when performing channel-level expansion with a ratio of 2, we expand the number of output channels in the first two convolutions $W^1$ and $W^2$, scaling them by 1/2. For the Last convolution $W^3$, we expand the number of input channels without any scaling. For simplicity, we omit the introduced noise $\epsilon$ introduced to the expanded weights. Since the output channel of the first two convolutions is expanded, we use $\{A, B\}$ to denote the concatenation of expanded channel features. Thus, the output of these three convolutions becomes can then be derived as follows:}
\\
{\small
\begin{flalign}
&\ Y_1 = \{W^1/2 \otimes X, W^1/2 \otimes X\}&
\end{flalign}
\begin{flalign*}
    Y_2 &= \{W^2/2 \otimes Y_1, W^2/2 \otimes Y_1\} \\
    &= \{W^2/2 \otimes \{W^1/2 \otimes X, W^1/2 \otimes X\}, W^2/2 \otimes \{W^1/2 \otimes X, W^1/2 \otimes X\}\} \tag{7}\\
    &= \{W^2/2 \otimes W^1/2 \otimes X + W^2/2 \otimes W^1/2 \otimes X, W^2/2 \otimes W^1/2 \otimes X + W^2/2 \otimes W^1/2 \otimes X\} \\
    &= \{W^2 \otimes W^1/2 \otimes X, W^2 \otimes W^1/2 \otimes X\} &
\end{flalign*}
\begin{flalign*}
    Y_3 &= W^3 \otimes Y_2 \\
        &= W^3 \otimes \{W^2 \otimes W^1/2 \otimes X, W^2 \otimes W^1/2 \otimes X\} \\
        &= W^3 \otimes W^2 \otimes W^1/2 \otimes X + W^3 \otimes W^2 \otimes W^1/2 \otimes X\} 
        \tag{8}\\
        &= W_3 \otimes W_2 \otimes W_1 \otimes X \\
        &= Y &
\end{flalign*}
}

\shulian{Clearly, given the same input $X$, the expanded model has exactly the same output as that of the original model, \emph{i.e.}, $Y_3 = Y$.}

\section{Parameter Sharing for Batch Normalization} \label{app:bn}
Batch normalization plays a crucial role in alleviating the vanishing gradient problem during the training of deep neural networks, ensuring stable convergence. However, since the student and dynamic teacher models may exhibit different data distributions during the training process, directly sharing the running statistics (\emph{i.e.}, running mean and running variance) of the batch normalization layers would lead to mutual interference between the two models, potentially causing training instability. To avoid this issue, we propose maintaining independent running means and variances ($\boldsymbol{\Tilde{\mu}}_s$, $\boldsymbol{\Tilde{\sigma}}^2_s$) and ($\boldsymbol{\Tilde{\mu}}_t$, $\boldsymbol{\Tilde{\sigma}}^2_t$) for the student and dynamic teacher models, rather than sharing the statistics.
\begin{figure}[h]
  \centering
  \includegraphics[width=0.6\linewidth]{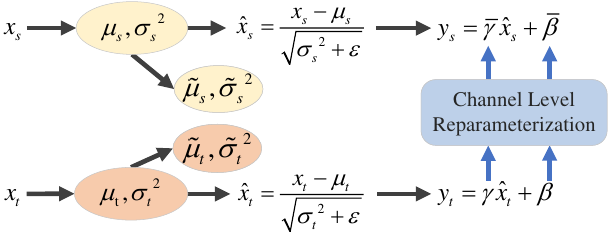}
  \caption{Illustration of parameter sharing for batch normalization in the student and dynamic teacher models. We main a separate set of running statistics for each model due to the distribution difference.}
  \label{fig:bn_param_sharing}
\end{figure}

As illustrated in Figure ~\ref{fig:bn_param_sharing}, each model independently calculates  its batch mean and batch variance from the input data, denoted as ($\boldsymbol{\mu}_s$, $\boldsymbol{\sigma}^2_s$) for the student model and ($\boldsymbol{\mu}_t$, $\boldsymbol{\sigma}^2_t$) for the dynamic teacher model. Subsequently, normalization is performed using these statistics, and the respective running statistics ($\boldsymbol{\Tilde{\mu}}_s$, $\boldsymbol{\Tilde{\sigma}}^2_s$) and ($\boldsymbol{\Tilde{\mu}}_t$, $\boldsymbol{\Tilde{\sigma}}^2_t$) are updated accordingly. The normalized outputs are then scaled and shifted using the learnable parameters $\boldsymbol{\gamma}$ and $\boldsymbol{\beta}$, which are shared across the two models through the channel ensemble strategy. This design effectively eliminates potential instabilities arising from sharing statistics, thereby guaranteeing the robustness of the entire training process.

\section{Experimental Settings}\label{app:setting}
\subsection{Distillation with A Static Teacher}\label{supp:experiment_setting1}
In this experiment, we adopt the standard data pre-processing pipeline, including random cropping, resizing to 224x224, random horizontal flipping, and normalization. By default, we employ the SGD optimizer with an initial learning rate of 0.1 and a momentum of 0.9. For convolutional neural networks, the batch size is set to 256 on 4 Nvidia Tesla V100 GPUs, while for vision transformers (ViTs), the batch size is set to 256 on 8 Nvidia Tesla V100 GPUs. The models are trained for 100 epochs with a learning rate decay factor of 0.1 applied every 30 epochs. The weight decay is set to 1e-4, and the weights for cross-entropy loss and KD loss are both set to 1.0.
For convolutional neural networks, we strictly follow the settings from~\cite{DBLP:conf/cvpr/ZhaoCSQL22, DBLP:conf/cvpr/Chen0ZJ21}. For the same architecture family, the teacher model is ResNet-34, and the student model is ResNet-18. For different architecture families, the teacher model is ResNet-50, and the student model is MobileNet-V1. Additionally, we explore the vision transformer architecture RVT ~\cite{mao2022towards}, employing RVT-S as the teacher model and RVT-Ti as the student model.

\subsection{Train from Scratch} \label{supp:experiment_setting2}
To further evaluate the efficacy of our approach, we conducted experiments without the reliance on a pre-trained static teacher model. Instead, we construct the dynamic teacher model via Inverse Reparameterization of the student model, as described in Sec.~\ref{sec:inverse_reparam}. Both the dynamic teacher and the student model are trained simultaneously from random initialization. During training, we leverage the additional loss terms introduced by our GPD, given by Eq.~\ref{eq: L_GPD}, to facilitate knowledge transfer from the dynamic teacher to the student model. We adopt the same data preprocessing and optimization strategies as described in Sec.~\ref{sec:distillation_with_ST}.

\subsection{Model Fine-Tuning} \label{supp:experiment_setting3}
In this part, we begin with pre-trained student models and aim to further improve their performance through our proposed approach. The pre-trained models are fine-tuned for 50 epochs, with the initial learning rate set to 0.1x the initial learning rate used during the pretraining stage. 
For the standard fine-tuning baseline, we fine-tuning the pre-trained models with cross entorpy loss.
Regarding our proposed GPD* method, we leverage the dynamic teacher model for distillation-based fine-tuning. Benefiting from Inverse Reparameterization, the dynamic teacher model initially exhibits the same accuracy as the student model. During the fine-tuning process, the dynamic teacher maintains a slightly higher accuracy than the student due to its increased capacity. We used the same loss function as in the experiment of training from scratch.

\section{Analysis of Computational Overhead in GPD} \label{app:overhead}
\shulian{
We analysis the computational overhead introduced by GPD. From Table~\ref{tab:computation_cost}, training a small student model (\emph{e.g.}, MobileNet) is not very efficient since it tends to come with a very low GPU utilization. Thanks to the high parallelism of GPU for large models (\emph{i.e.}, our expanded dynamic teacher), GPD only introduces 33\% extra overhead while yielding a performance improvement of 1.58\%. Furthermore, as shown in Table~\ref{tab:sota_comparison}, our improvement is significantly larger than recent methods, achieving a good trade-off between performance and training cost. Moreover, since the static teacher used for distillation is often very large with high computational cost, the most important goal of distillation should be improving the performance by sacrificing the training efficiency. From this point of view, our GPD could be a very strong baseline for distillation.}

\begin{table}[h]
\centering
\caption{Comparison of computation cost on the experiment of ResNet50 → MobileNet distillation. We measure the training time on 4 A100 GPUs with a batch size of 512 on ImageNet.}
\label{tab:computation_cost}
\begin{tabular}{lcc}
\toprule
Method & Acc. (\%) & Training Time per Epoch (min) \\
\midrule
DKD & 72.05 & 12 \\
DKD + GPD & 73.63 (+1.58) & 16 (+33\%) \\
\bottomrule
\end{tabular}
\end{table}

\end{document}